# Using LSTM and GRU With a New Dataset for Named Entity Recognition in the Arabic Language

Alaa Shaker[1], Alaa Aldarf[1] and Igor Alexandrovich Bessmertny[1]

[1]ITMO University, Kronverksky Pr. 49, bldg. A, St. Petersburg, 197101, Russian Federation

### Abstract

Named entity recognition (NER) is a natural language processing task (NLP), which aims to identify named entities and classify them like person, location, organization, etc. In the Arabic language, we can find a considerable size of unstructured data, and it needs to different preprocessing tool than languages like (English, Russian, German...). From this point, we can note the importance of building a new structured dataset to solve the lack of structured data. In this work, we use the BIOES format to tag the word, which allows us to handle the nested name entity that consists of more than one sentence and define the start and the end of the name. The dataset consists of more than thirty-six thousand records. In addition, this work proposes long short-term memory (LSTM) units and Gated Recurrent Units (GRU) for building the named entity recognition model in the Arabic language. The models give an approximately good result (80%) because LSTM and GRU models can find the relationships between the words of the sentence. Also, use a new library from Google, which is Trax and platform Colab.

## 1. Introduction

Named entity recognition (NER), as known it is one of the natural language tasks (NLP) that try to extract information from text by identifying the names of entities [1]. Identifying the named entities can help in many tasks, for example, information retrieval, answer questions, machine translation, automatic text summarization etc.[2] Arabic NER has gained considerable interest and focus from the researchers due to the popularity of the Arabic language, as it is the native language for more than 325 million people living in Arab countries. However, there is a lack of structured databases for NER task in the Arabic language, and this work tries to give an Arabic dataset for this tasks in the areas (Geography, History, Medical, Sport, Technology, News and Cooking). Also, there are many differences between the Arabic language and other languages [3]. First of all, it is written from right to left. The second difference is (تَشْكِيل) in English (Tashkil), the literal meaning of Tashkil is (forming). In some Arabic countries use the word (الحركات) in English (Alharakat) instead of word tashkil.

We can understand them as short vowels, and there are five types of Tashkil (kasrah, dammah, fatha, Sukun, shadah). The third difference that the Arabic language has (التنوين) in English (Altanween) or (nunation), which has three types (Alrafa, Alnasb, Aldam). Only nouns get nunation, but not all nouns. Also, it is rare to use, especially in the digital text (often the writers forget to add them to text), so we cannot depend on them to identify the nouns. Therefore, we need to be careful when processing Arabic texts because what appears to computer perception as a difference, in reality, is the same for human conception. For example, if we take the two words (العيون) and (العيُون, العيْون ). Both words have the same meaning, "eyes," but different digital codes because of Tashkil's existence, so we provide the needed function to normalize the Arabic words to solve this problem. In addition to this, in the Arabic language, there is no upper/lower case for the first letter in the name like English, Russian, etc. Researchers use one of three main approaches for building NER systems. As described in [4, 5], these approaches are linguistic-rule-based, machine-learningbased, and hybrid-based approaches. Rule-based approach: when the researchers create the rule or template to extract the names of entities from input sentences, in 1995 rule-based NER system was developed that used specialized dictionaries of names that included countries, names of cities, etc [6]. One of the significant challenges of the rule-based NER systems is that the researchers must have sufficient knowledge and understanding of language grammar. On the other side, machine language systems (ML systems) do not need to know language grammar but need a massive dataset to train a model. There are two types of ML NER systems:

Unsupervised ML – there is no need for linguists to work on training or testing sets and is mainly limited to clustering. Another type is the supervised ML – linguists need to build the training and test sets, which consist of texts and labels, but if we have enough NER datasets, there is no need to know the language's grammar. Some examples of the ML techniques being used for NER algorithms cover Support Vector Machine, the artificial neural network (ANN), Hidden Markov Model (HMM), and Decision Trees, etc [7]. Deep learning is considered a subdomain of ML; the model of deep learning consists of many layers. There are two main types of layers: convolutional neural networks (CNN)-based and recurrent neural networks (RNN)-based, and the RNN are designed to work with sequential data. In this work, we use Gated Recurrent Unit (GRU) and Long Short Term Memory (LSTM) [8] in two models to make a comparison between their results and to check the efficiency of our dataset. GRU and LSTM were founded to solve the vanishing gradient problem. This problem existed because the traditional RNN cannot hold the relationship between the words in first places and the words in last places in long sentences. This work consists of the following sections: the first section talks about the dataset, the second one talks about the used model, the third section discusses the result, and the last one is the conclusion.

## 2. Dataset

In an overview of the existence NER dataset in the Arabic language, we can note that all datasets cover the topics of politics, economics, more recently, the medical field. Therefore, the trained model is limited to identifying the names of entities in the previously mentioned fields. This work provides a new dataset in seven different fields (Geography, History, Medical, Sport, Technology, News, and Cooking) to achieve variety, which gives the trained model on this dataset the ability to find the entities' names in different domains. Beginning, inside, End, Single, Out-side (BIOES) format [9] is used to label words in the dataset, which gives the ability to label nested names and their beginning/end. The BIOES format is better than the IOB format that does not have this ability. Especially in a language like the Arabic language, because there are a lot of nested names. For example, " محمد بن اسماعيل بن جعفر الصادق " in English "Muhammad bin Ismail bin Jaafar al-Sadiq" this name consists from six-words and its label in BIOS as "B-PER I-PER I-PER I-PER I-PER E-PER". The entities' names were labeled in nine categories: Person (PER), Location (LOC), geopolitical (GEO), time (TIM), profession (PRO), organization (ORG), disease (DIS), geography (GEO), and miscellaneous (MISC). "Figure 1" shows how data is organized in excel files.

| file_name | sentence | word | tag |
|---|---|---|---|
| 31 | 1 | قامت | O |
| 31 | 1 | في | O |
| 31 | 1 | مدينة | B-LOC |
| 31 | 1 | اشور | E-LOC |
| 31 | 1 | في | O |
| 31 | 1 | شمال | B-LOC |
| 31 | 1 | بلاد | I-LOC |
| 31 | 1 | ما | I-LOC |
| 31 | 1 | بين | I-LOC |
| 31 | 1 | النهرين | E-LOC |

**Figure 1:** The dataset structure.

## 3. The proposed named entity recognition model

### 3.1. Preprocessing the data

In the other NER systems of languages like English, there is no need to preprocess the names' entities because no changes happened to names. But in the Arabic language same word is written in different ways [10] because, in the Arabic language, there are Altashkil and Altanween, as we mentioned before. For example, the two words "العيونَ" and "العيونُ", "العيون" they have the same meaning (in English "Eyes") but different code because of the presence of the Altashkil, which makes NER systems unable to label them properly. So we need to remove "Altashkil" and "Altanween" to allow the system to understand both words as equivalent; this operation is called normalization

### 3.2. The model

In our model, we will test two different cells LSTM and GRU. LSTM was first introduced in 1996 [8] to solve the problems of vanishing gradient [11] and gradient explosion [12] that existed in ordinary RNN cells. In the LSTM, there are three gates (input/update gate, forget gate, output gate). The computational operations of the LSTM cell consist of seven equitations described by the equitations (1, 2) [13]. The internal architecture of the LSTM unit is shown in "figure2".

$$i_t = \sigma(W_i X_t + R_i h_{t-1} + b_i)$$
$$f_t = \sigma(W_f X_t + R_f h_{t-1} + b_f)$$
$$o_t = \sigma(W_o X_t + R_o h_{t-1} + b_o) \quad (1)$$
$$C_t' = \tanh(W_c X_t + R_c h_{t-1} + b_c)$$
$$C_t = f_t \odot C_{t-1} + i_t \odot C_t'$$
$$h_t = o_t \odot \tanh(C_t)$$

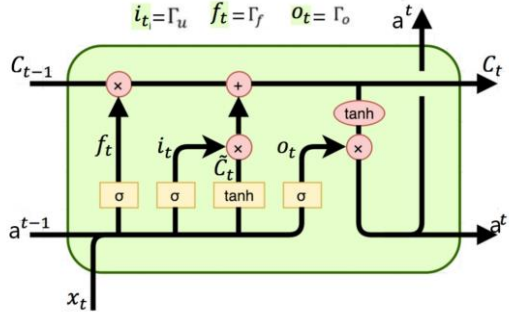

Figure 2: The internal architecture of the LSTM unit. (https://tinyurl.com/LSTM-VS-GRU/)

On the other hand, the internal structure of GRU is shown in "figure 3". We can note that GRU consists of two gates (reset gate, and update gate). The GRU's unit requires just five arithmetic operations and has two gates; therefore, it is less complex than LSTM.

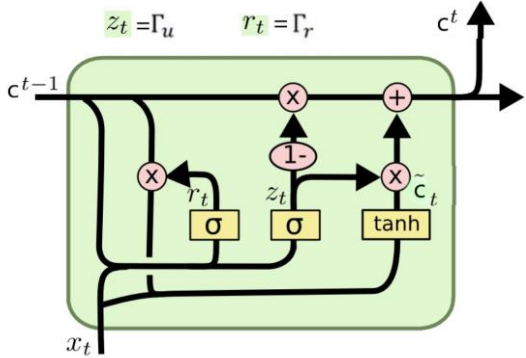

Figure 3: The internal architecture of the GRU unit. (https://tinyurl.com/LSTM-VS-GRU/)

The equitation (2) explain the computation operations within the GRU unit

$$r_t = \sigma(W_{rx} X_t + W_{rc} C_{t-1} + b_r)$$
$$z_t = \sigma(W_{zx} X_t + W_{zc} C_{t-1} + b_z) \quad (2)$$
$$C_t' = \tanh(W_{nx} X_t + W_{nc}(r_t \odot C_{t-1}) + b_n)$$
$$C_t = (1 - z_t) \odot C_t' + z_t \odot C_{t-1}$$

After explaining how the LSTM and GRU units work, which form the primary cells to build our deep learning model for named entity recognition, we need to put these units in the architecture to build the model. "Figure 4" shows the proposed model.

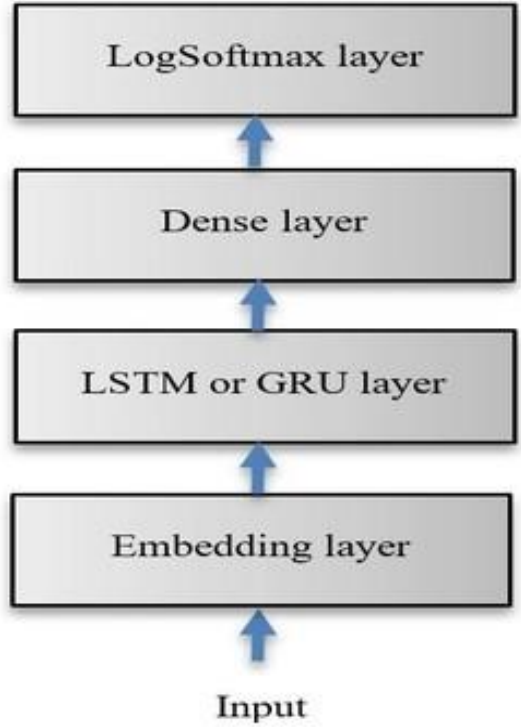

Figure 4: The NER model architecture.

The first layer in the model is the embedding layer that converts the word into a vector. In this model, each input word is converted into a vector with fifty elements. Secondly, the LSTM/GRU layer, the third layer, is a dense layer that applies the Relu activations functions, and the last layer is implemented with Logsoftmax.

## 4. Results

The first result of this work is the new dataset for the Arabic NER task. The dataset consists of three main files (training, validation, and test); each file contains

**Table 1**

Statistical information about the database used

|  | Number of files | Number of sentence | Number of words |
|---|---|---|---|
| Training set | 120 | 1731 | 31586 |
| Validation set | 15 | 218 | 3107 |
| Test set | 8 | 155 | 1687 |
| The total | 143 | 2104 | 36380 |

**Table 2**

The number of words belonging to each of nine categories

|  | PER | GPE | LOC | ORG | TIM | PRO | MISC | DIS | GEO |
|---|---|---|---|---|---|---|---|---|---|
| Training | 991 | 2091 | 941 | 494 | 1088 | 388 | 2524 | 604 | 576 |
| Validation | 83 | 124 | 61 | 51 | 120 | 48 | 215 | 73 | 90 |
| Test | 38 | 75 | 35 | 40 | 63 | 45 | 154 | 34 | 62 |
| Total | 1112 | 2290 | 1037 | 585 | 1271 | 481 | 2893 | 711 | 728 |

**Table 3**

The accuracy of models on validation and test sets

| iterations | The model | Accuracy in validation step | Accuracy in test step |
|---|---|---|---|
| 500 | LSTM NER model | 81.86% | 80.24% |
| 500 | GRU NER model | 78.86% | 77.78% |
| 1000 | LSTM NER model | 79.88% | 80.06% |
| 1000 | GRU NER model | 76.26% | 75.38% |

**Table 4**

The time for training models within different iterations

| iterations | The model | Time needed in second |
|---|---|---|
| 500 | LSTM NER model | 88.003 |
| 500 | GRU NER model | 89.787 |
| 1000 | LSTM NER model | 103.98 |
| 1000 | GRU NER model | 113.829 |

seven files because the dataset covers the seven fields mentioned above. These files contain one hundred fortythree excel files. Each excel file contains four columns (file_name, sentence, word, tag). So we can use this dataset to train models for tasks like Arabic NER, classify texts, and classify sentences. Table 1 shows the statistical numbers of files, number of sentences, and number of words included in the dataset.

The second point is the NER model. The environment used for the experiment is COLAB from Google, which allows anyone to write and execute Python code through the browser. Also, to implement the deep learning model, the Trax library was used. Trax is an end-to-end library for deep learning from the Google Brain team. This article discusses the NER task, a multi-class classification, so cross-entropy is selected for the loss function.

The chosen optimizer is the Adam optimizer [14], with an alpha parameter equal to 0.01, suitable for large-scale corpus [14]. The Embedding layer dimension is (vocabulary_size,50), so each word from text is converted to a vector from fifty elements, and for the LSTM/GRU layer, the dimension is fifty units. The Dense layer dimension equals the number of names' entities' categories, which in this dataset is thirty-seven classes. The size of the last layer Logsoftmax should be equal to the previous layer, so it is also thirty-seven units. The accuracy measure used in this work is precision, which equals dividing the accurately predicted labels by the total number of labels in the sentence. The number of trainable parameters in LSTM equals 608937, and in the GRU model equals 603887.

Table 3 shows that the LSTM model gives the best result for (500, 1000) iterations, which equals 80% for both validation and test dataset, which is relatively good results. The GRU model gives well results but less than the LSTM model. We can note from the Table 3 that more times iterations does not mean better results. That is because the model reached the threshold after which it cannot increase the accuracy, and the accuracy remains around it. In addition, the results show that the trained models can generalize their knowledge, and they can successfully find the named entities that exist in files of validation and test set.

## 5. Conclusion

This article introduced an Arabic NER dataset that is diverse and includes nine classes of named entities. The LSTM and GRU models' results show that after training models, they give a good result on validation and test, which means that models can generalize their knowledge and successfully detect the named entities within validation and test sets. The texts are variety in seven different domains. Both models (LSTM, GRU) give good results, and they can find the entities' names with a precision of approximately (80%). In the future, the dataset will be increased by adding more labeled texts, and we will try to use it with other deep learning techniques to get better accuracy.

## A. Online Resources

The links below contain a link to the Trax library documentation sources and the Git hub repository. In the repository, you can find the dataset and the codes to read and preprocess the data.

- Trax library documentation,
- GitHub.